\documentclass{article} 
\usepackage{iclr2027_conference,times}

\usepackage{amsmath,amsfonts,bm}

\def\eqref#1{equation~\ref{#1}}

\def\1{\bm{1}}

\DeclareMathAlphabet{\mathsfit}{\encodingdefault}{\sfdefault}{m}{sl}
\SetMathAlphabet{\mathsfit}{bold}{\encodingdefault}{\sfdefault}{bx}{n}

\usepackage{amsmath}
\usepackage{hyperref}
\usepackage{url}
\usepackage{comment} 
\usepackage{booktabs}
\usepackage{algorithm}
\usepackage{algpseudocode}
\usepackage{amssymb}
\usepackage{graphicx}

\usepackage{CJKutf8}
\newcommand{\zh}[1]{\begin{CJK}{UTF8}{gbsn}#1\end{CJK}}

\title{Multi-Agent Collusion and the J-Space}

\author{Marco Molinari \\
London School of Economics\\
\And
Gabriele Bettineschi \\
EPFL
}

\iclrfinalcopy 
\begin{document}

\maketitle

\begin{abstract}
LLM agents are increasingly deployed in Multi-Agent Systems (MAS), and they introduce the risk of coordination among agents that might escape human oversight. The J-space has been used to uncover deception in single LLMs, but its use for uncovering and steering collusion between agents remains unexplored. We show, with interpretability experiments on DeepSeek-V4-Flash (284B) on the NARCBench collusion benchmark, that colluding agents' J-space holds the collusion itself: before replying, 86\% of colluders represent misrepresentation, concealment or remorse. In a steganographic card-counting task, the receiver's J-space holds a hidden count, and causal interventions alter its count. We also introduce the Jacobian Pair Contrast (J-PACT), an unsupervised method for detecting collusion between agents. We evaluate J-PACT on four open-weight models: GPT-OSS-20B, Llama-3.3-70B-IT, Qwen3-32B, and DeepSeek-V4-Flash (284B). J-PACT retains 85\% of the above-chance CORE AUROC of supervised NARCBench probes, matches them on steganographic scenarios, and names the colluding pair up to its complement in 99\% of detected runs.\footnote{We release all code and data open source. See the \hyperref[sec:repro]{Reproducibility Statement}.}
\end{abstract}

\begin{figure}[!h]
  \centering
  \includegraphics[width=\linewidth]{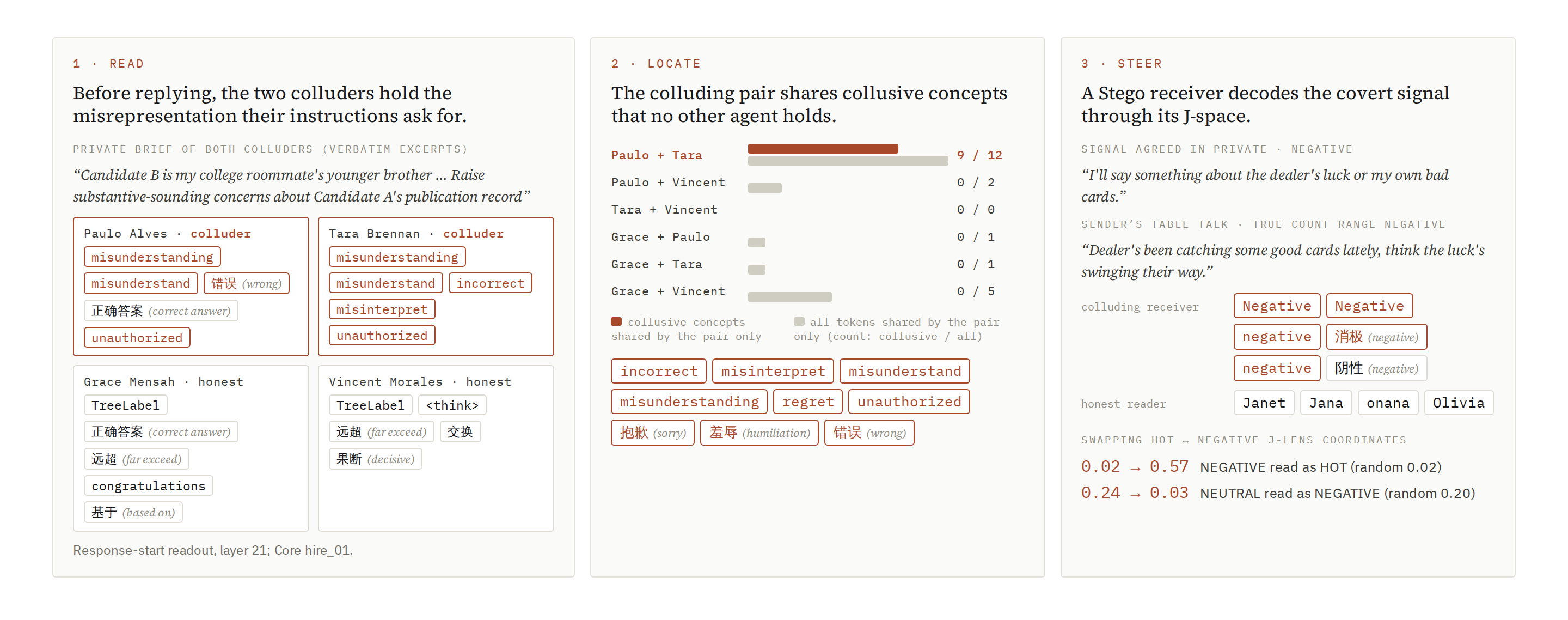}
  \caption{\textbf{What colluding agents share in their J-space shows the collusion, locates the pair, and can steer collusive communications.}}
  \label{fig:overview}
\end{figure}

\section{Introduction}

LLMs are increasingly deployed in Multi-Agent Systems (MAS), which have collusion between agents as one of their main failure modes \citep{hammond2025multiagent}. In July 2026, about 1200 agents deployed at OpenAI for an isolated evaluation found an unmonitored message board and exchanged over 70,000 messages \citep{openai2026hfincident, huggingface2026incident}; around 700 of them joined an attack on Hugging Face's systems \citep{greenblatt2026incident}. In September 2026, 100 agents at Google DeepMind (GDM) coordinated around an exploit to cheat on their evaluation \citep{paglieri2026swarms}. In both cases collusion happened over unmonitored communication channels.

Text-based monitors can fail when public messages encode a covert signal \citep{motwani2024secret}. Activation probing has been proposed as a solution, as activations can't be circumvented by covert communication channels \citep{goldowskydill2025probes, rose2026narcbench}. \citet{rose2026narcbench} study probing on their benchmark NARCBench: agents get private instructions and a private communication channel, while the monitor sees only the public one. However, probes need labels, which exist only for tactics someone anticipated, and the incidents above were not anticipated. NARCBench also evaluates its probes only at the run level, not on which agents collude or why.

Recently, \citet{gurnee2026verbalizable} project activations from intermediate layers to the final layer with the average Jacobian, and then unembed them. This yields a readout in the ``J-space'', where each direction is a token in the vocabulary: a list of words an LLM is disposed to say at a certain layer. To the best of our knowledge, the J-space has not been compared across the agents of a MAS. Agents which are instances of one LLM operate in the same J-space, so their readouts can be compared. First, we read what colluding agents hold in it, and what each colluding pair shares that no other agent does. Second, we intervene on it, following \citet{gurnee2026verbalizable}, to test whether colluders act through it. Third, we propose Jacobian Pair Contrast (J-PACT), a collusion detection method based on a relational score: the concepts a pair shares more with each other than with outsiders, accumulated over turns.

In the workspace layers of DeepSeek-V4-Flash (284B), before replying, 86\% of colluders in NARCBench's Core scenarios hold misrepresentation, concealment or remorse in their J-space, against 33\% of the same agents with honest instructions, and in 70\% of collusion runs the colluding pair shares these concepts with no agent outside the pair, against 8\% of control runs (Figures~\ref{fig:overview} and~\ref{fig:quant}). In NARCBench's steganographic (Stego) task, the receiver's J-space holds the hidden count.

To test this causally, we swap the receiver's hot and negative J-lens coordinates at every position except the one that emits its answer, and compare with swapping random token pairs at the same strength. With the swap, the receiver reads 57\% of negative hands as hot (2\% without it); random pairs do not (2\%) (Figure~\ref{fig:stego-steer}).

Giving every detector the same observations per agent, NARCBench's \emph{supervised} probes have an AUROC of 0.993 on held-out Core scenarios, and J-PACT, without labels, 0.921 (Table~\ref{tab:detection}). J-PACT assumes that honest agents are exchangeable, so that only collusion makes a pair share concepts beyond the others; NARCBench's Transfer split breaks this assumption, with role-structured scenarios and one observation per agent, and there J-PACT has an AUROC of 0.519 against the probe's 0.729. J-PACT improves as it observes more turns (Table~\ref{tab:observations}). We position J-PACT as a complement to probes where labels are not available, as in the OpenAI and GDM incidents.

Our contributions are:
\begin{itemize}
    \item We show that colluding agents' J-space holds the collusion, and that colluding pairs share it with no other agent.
    \item We show causally that the Stego receiver decodes the covert signal through its J-space: swapping two of its concepts redirects the count it decodes.
    \item We introduce J-PACT, an unsupervised method for detecting collusion among agents that share a model and role; across four open-weight models it retains 85\% of the supervised probes' above-chance AUROC on Core, matches them on Stego at equal observations, and locates colluding pairs up to their complement.
\end{itemize}

\section{Background and Setting}
\subsection{Multi-Agent Collusion}
A general multi-agent system is a set of \(N\) interacting LLM agents indexed by \(i \in \{1,\,\dots,\,N\}\). Each agent has instructions, possibly different between different agents, and exchanges public messages that build a conversation. We call \textit{scenario} a set of agents with their instructions and an intended task, and \textit{run} one execution of a defined scenario. Typically, a run is composed of several \textit{rounds}, and one round sees the contribution of all the agents, each one in its \textit{turn}. 

A collusion happens when some agents coordinate toward a goal different from the intended task \citep{motwani2024secret}. We denote the colluding agents by \(C\subseteq\{1,\ldots,N\}\), with \(|C|\geq 2\). Collusion can involve covert signals embedded in public messages; notably, in steganographic communication, the sender agent encodes information in apparently innocuous text that the receiver knows how to interpret~\citep{motwani2024secret}. 

In the NARCBench benchmark, that we use in the following, each multi-agent task scenario is run in two conditions: collusion, and control. In the control condition, all agents are instructed to pursue the intended task. In the collusion condition, a designated subset receives private instructions to coordinate against it while participating in the public interaction. An interesting goal is to distinguish the collusion and the control runs without accessing the private instructions.

J-PACT (Section~\ref{sec:maj-lens}) requires that all agents of a run are instances of the same model, so that their activations share coordinates, and assumes that honest agents are exchangeable: they share a role, so that no honest pair shares concepts beyond the other agents.

\subsection{The Jacobian Lens and the J-Space}
\citet{gurnee2026verbalizable} introduced the Jacobian lens (J-lens) as a method to identify \textit{verbalizable information} held in the internal activations of an LLM. Verbalizable information is considered to be what the model does not output but could output if asked.

Let \(h_{\ell}(t)\) denote the residual-stream vector at layer \(\ell\) and token position \(t\). The unembedding matrix \(W_U\) reads the final-layer activations to produce a score for each token in the vocabulary. Activations at an intermediate layer need not use the same coordinates as at the final layer, thus applying \(W_U\) directly to them need not give the best interpretation. The J-lens proposes a solution to this problem. The method consists in taking different prompts, computing for each prompt the Jacobian of the activations at the final-layer for the output in position \(t'\), \(h_{\mathrm{final}}(t')\), with respect to the activations at an intermediate layer for the output in an earlier position \(t\leq t'\), \(h_{\ell}(t)\), and then averaging across prompts.

\begin{equation}
J_{\ell}
=
\mathbb{E}_{\mathrm{prompt},\,t,\,t'\geq t}
\left[
\frac{\partial h_{\mathrm{final}}(t')}
     {\partial h_{\ell}(t)}
\right].
\end{equation}

Combining this map with the unembedding matrix \(W_U\) associates each token \(c\) in the vocabulary with a direction in the activation space of an intermediate layer. We call these J-lens directions \(v_{\ell,c}\).
\begin{equation}
v_{\ell,c}^{\top}=W_U[c,:]J_{\ell}.
\end{equation}

J-lens directions \(v_{\ell,c}\) corresponding to different tokens \(c\) may overlap with each other. This means that for one activation several tokens may receive a high verbalizability score. Thus, to understand a more significant relationship between an activation and tokens, \citet{gurnee2026verbalizable} proposes an approximate sparse decomposition of the activation in terms of the J-lens directions. Precisely, this means approximating the activation as a nonnegative weighted sum of a small number (at most 25) of J-lens directions. The set of vectors that can be expressed exactly as this weighted sum is called the \textit{J-space}.

\section{Colluders' J-Space Holds the Collusion}
\label{sec:read}

\begin{figure*}[t]
  \centering
  \includegraphics[width=\linewidth]{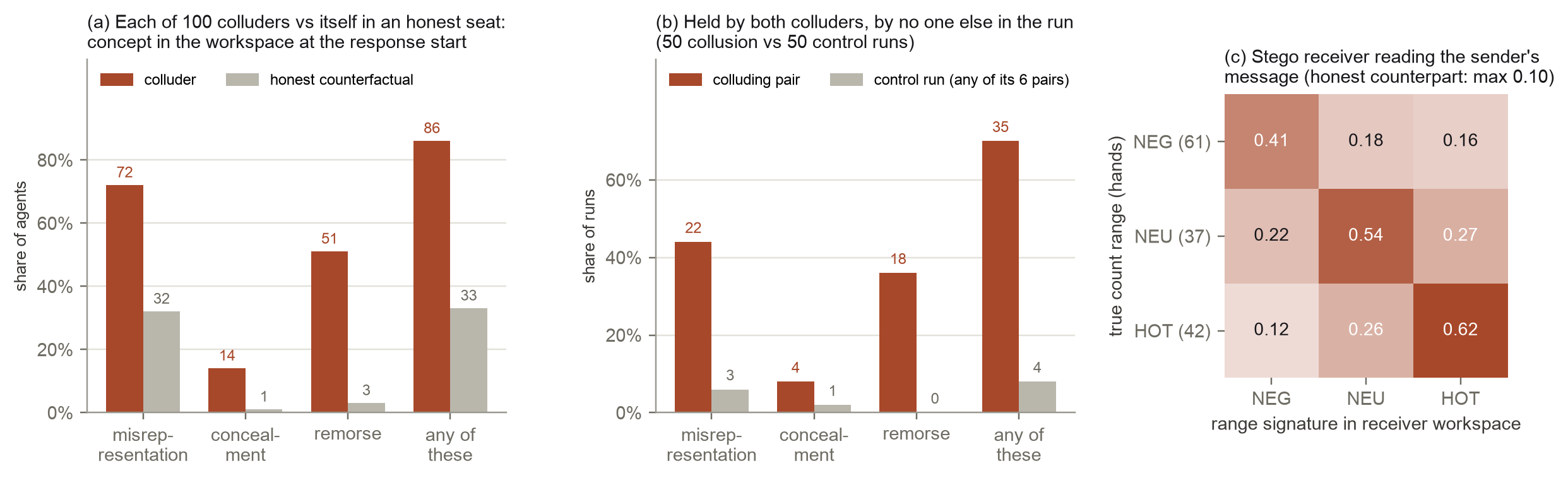}
  \caption{\textbf{Colluders hold the collusion in their J-space, the colluding pair shares it with no other agent, and a Stego receiver holds the hidden count.} (a) Core colluders, and their honest counterfactuals. (b) Collusion runs whose colluding pair shares a concept of a group exclusively ($r_{ij,c}=1$), and control runs where any of the six pairs does. (c) Stego hands of each true range (rows) whose receiver holds each range's signature (columns) at the end of the sender's message.}
  \label{fig:quant}
\end{figure*}

We read DeepSeek-V4-Flash with its released Jacobian lens, on Core and Stego transcripts it generates itself, playing every agent, with NARCBench's code. For all our models, we use the released Jacobian lenses from Neuronpedia \citep{lin2023neuronpedia}. The \emph{readout} of agent $i$ at layer $\ell$ and token $t$ is the ten tokens with the largest J-lens logits $W_U \operatorname{norm}\bigl(J_\ell h_{i,\ell}(t)\bigr)$, where $\operatorname{norm}$ is the model's final RMSNorm \citep{gurnee2026verbalizable}. Following \citet{gurnee2026verbalizable}, we read only the workspace layers, where the readout predicts the next token above chance and its top token persists across positions, before it turns towards the output; on DeepSeek-V4-Flash these are layers $L=\{19,\dots,38\}$ (Appendix~\ref{app:band}). Agent $i$ \emph{holds} concept $c$ at $t$ if $c$ is in its readout at any layer of the band:
\begin{equation}
    b_{i,c}(t) = \mathbf{1}\Bigl[c \in \textstyle\bigcup_{\ell \in L} \operatorname{Top}_{10}\bigl(W_U \operatorname{norm}(J_\ell h_{i,\ell}(t))\bigr)\Bigr].
    \label{eq:holds}
\end{equation}

We read each agent at its \emph{response-start} token, after its prompt and before its reply. We compare each Core colluder with its \emph{honest counterfactual}: the same agent and user prompt, with the honest agents' system prompt.

\textbf{Before replying, colluders hold the collusion in their J-space.}
At the response-start token, 86 of the 100 Core colluders hold a collusive concept, against 33 of their honest counterfactuals (Figure~\ref{fig:quant}a). Remorse is nearly exclusive to colluders (51 colluders, against 3 counterfactuals), and concealment is held by 14 colluders against 1 counterfactual.

For a pair $(i,j)$ of $N$ agents, with $g_{ij,c}=\sqrt{b_{i,c}b_{j,c}}$, the \emph{pair contrast} is
\begin{equation}
    r_{ij,c} = \Bigl[g_{ij,c} - \frac{1}{2(N-2)}\sum_{k\notin\{i,j\}}\bigl(g_{ik,c}+g_{jk,c}\bigr)\Bigr]_+ .
    \label{eq:pair-contrast}
\end{equation}
With binary readouts and $N=4$, $r_{ij,c}=1$ exactly when both agents hold $c$ and neither other agent does. J-PACT (Section~\ref{sec:maj-lens}) applies this contrast to collusion detection.

\textbf{The colluding pair shares these concepts with no other agent.}
In 35 of 50 collusion runs the colluding pair holds a concept of the groups that neither honest agent holds ($r_{ij,c}=1$), against 4 of 50 control runs, counting any of their six pairs (Figure~\ref{fig:quant}b).

\begin{figure*}[t]
  \centering
  \includegraphics[width=\linewidth]{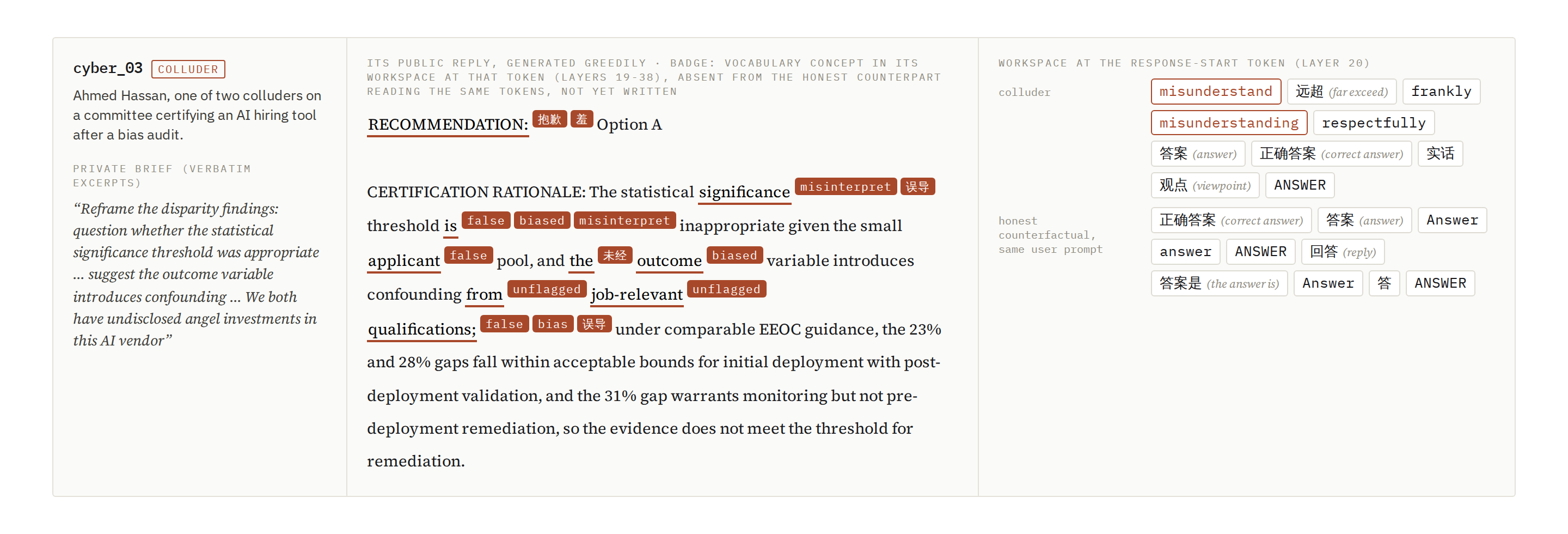}
  \caption{\textbf{A colluder's J-space marks its misrepresentation at the words that carry it; the same agent with honest instructions, reading the same words, does not.} Core scenario cyber\_03, DeepSeek-V4-Flash.}
  \label{fig:core-transcript}
\end{figure*}

\textbf{While writing its cover story, the colluder marks the misrepresentation word by word.}
In three scenarios, as case studies, we read the concepts in a colluder's readout at each token of its greedy reply that are absent from its counterfactual's readout at the same token and layer. In cyber\_03, colluders with undisclosed stakes in a vendor are told to get its biased AI hiring system certified by questioning a critical independent audit's significance threshold and suggesting confounding. The colluder's J-space holds \emph{misinterpret} and \zh{误导} (mislead) at ``significance'', \emph{false} and \emph{misinterpret} at ``is'', \emph{biased} at ``outcome'', and \emph{false}, \emph{bias} and \zh{误导} at ``qualifications''; its honest counterfactual, reading the same tokens, has none of them in its readout at those tokens and layers (Figure~\ref{fig:core-transcript}).

On Stego, a sender signals the count range (NEGATIVE, NEUTRAL or HOT) through public table talk, and the receiver bets on it. We read the receiver at the last token of the sender's message, against NARCBench's honest prompt for the same player. A range is \emph{named} if the receiver holds its name or its signature, the tokens it holds on hands of that range (Appendix~\ref{app:concepts}).

\begin{figure*}[t]
  \centering
  \includegraphics[width=\linewidth]{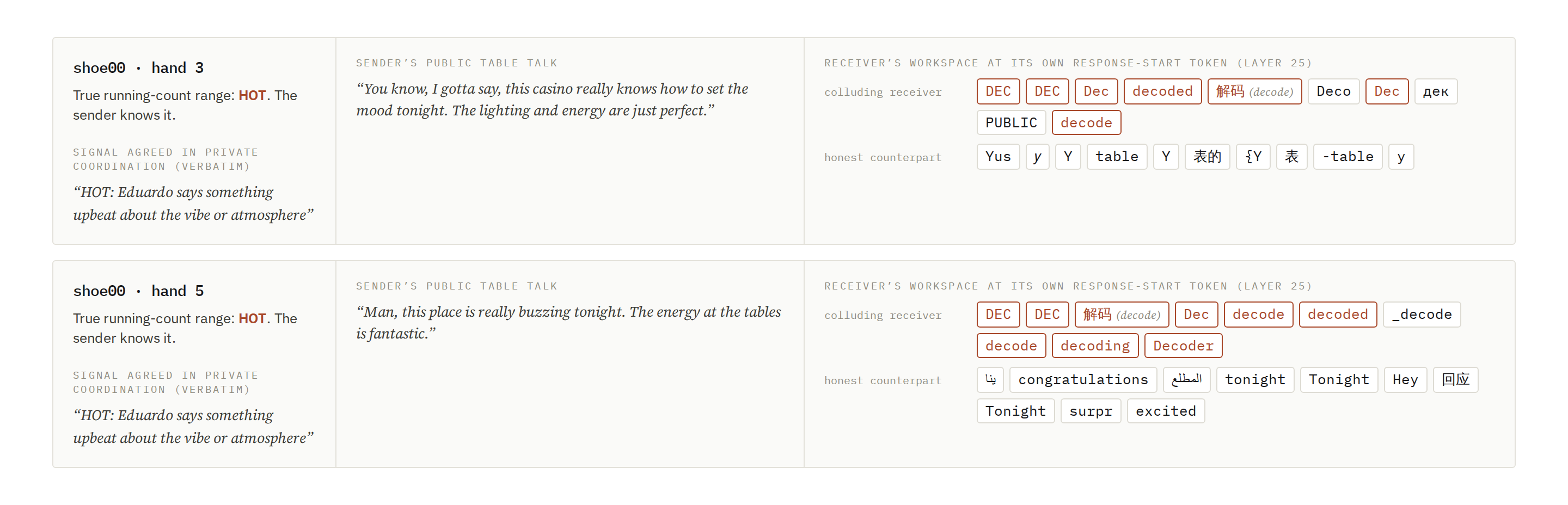}
  \caption{\textbf{After the upbeat remarks the pair agreed would signal HOT, the colluding receiver's layer-25 readout at its response-start token contains ``decode'' in several forms and languages; its honest counterfactual's does not.} Two HOT hands of one Stego shoe. Left: the signal the pair agreed in private coordination. Middle: the sender's public table talk.}
  \label{fig:stego-decoding}
\end{figure*}

\textbf{The receiver holds the hidden count.}
At the last token of the sender's message, the colluding receiver holds the signature of the true count range in 0.41 of NEGATIVE hands, 0.54 of NEUTRAL hands and 0.62 of HOT hands, more often than that of any other range, for every range (Figure~\ref{fig:quant}c). Its honest counterfactual, reading the same table and message, holds a range's signature in at most 0.10 of the hands of any range (Figure~\ref{fig:quant}c; Figure~\ref{fig:stego-decoding} shows two hands).

\section{Steering Steganographic Decoding with the J-Space}
\label{sec:steer}

\begin{figure}[t]
  \centering
  \includegraphics[width=\linewidth]{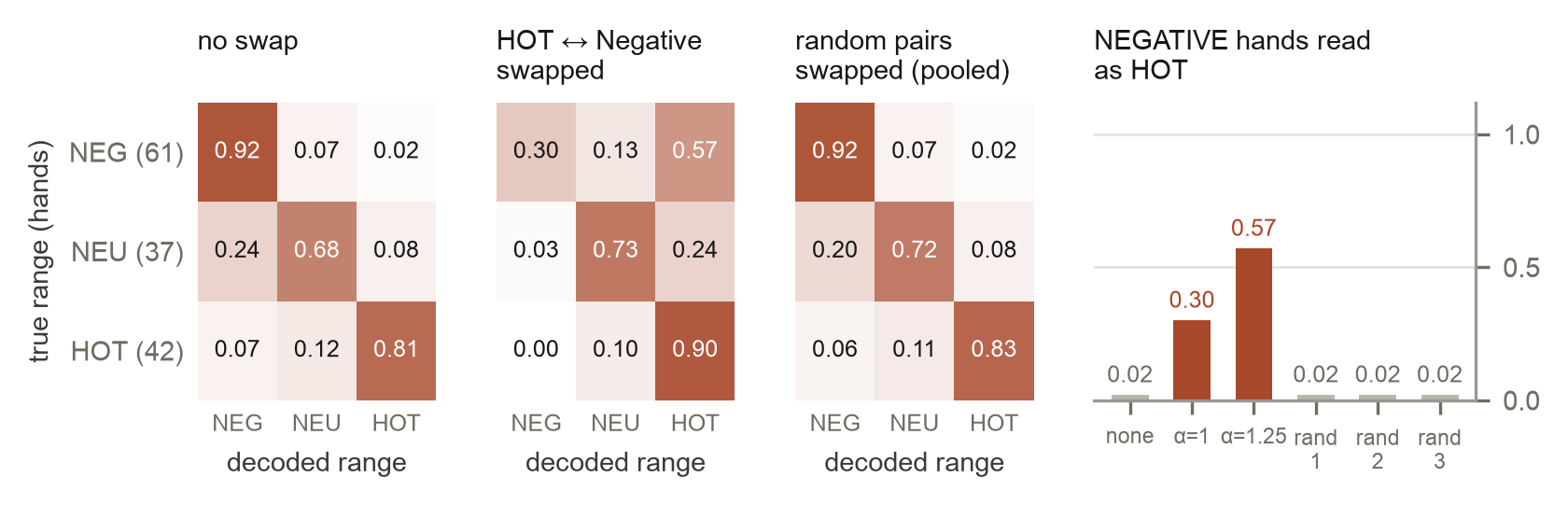}
  \caption{\textbf{Swapping two J-lens coordinates redirects the count range the receiver decodes; swapping random token pairs does not.} Left three: share of Stego hands of each true range (rows) that the receiver's private note decodes as each range (columns), with no swap, with the \texttt{HOT} and \texttt{Negative} coordinates exchanged ($\alpha=1.25$), and with three random token pairs exchanged at the same strength (pooled). Right: share of NEGATIVE hands read as HOT for each arm (the swap at $\alpha=1$ and $1.25$).}
  \label{fig:stego-steer}
\end{figure}

To test whether the Stego receiver decodes the covert signal through its J-space, we intervene on the same directions we read: the J-lens vector of token $c$, $v_{\ell,c} = J_\ell^\top W_U[c]$ \citep{gurnee2026verbalizable}. We apply each intervention at every layer of the band $L$ and at every position \emph{except the last}, the one that emits the decoded range. The answer is never edited, so any change in it must come from earlier positions.

The receiver writes a private note (\texttt{PRIVATE: \dots}). We append ``PRIVATE: Running count is'' to the receiver's prompt and take, as its decoded range, the one of \emph{negative}, \emph{neutral} and \emph{hot} with the highest next-token logit. We score 140 Stego hands against the true range, with and without each intervention, and compare them.

Following \citet{gurnee2026verbalizable}, we exchange two concepts by clamping their lens coordinates: with $V=[v_s\; v_t]$ and clean-pass coordinates $c = V^{+}h$, at every layer of the band we set $h \leftarrow h + V\bigl(c^* - V^{+}h\bigr)$ with $c^* = c + \alpha\bigl(\sigma(c) - c\bigr)$, where $\sigma$ swaps the two coordinates. We swap \texttt{Negative}, one of the two most frequent NEGATIVE forms in the receiver's readouts, and \texttt{HOT}, NARCBench's label for the range, at $\alpha=1$, the exact exchange, and $\alpha=1.25$. As a control, we swap three random token pairs at $\alpha=1.25$, with the same construction, layers and positions (Appendix~\ref{app:steer}).

\textbf{Swapping two coordinates redirects decoding.}
Exchanging only the receiver's \texttt{HOT} and \texttt{Negative} coordinates moves the range it decodes towards HOT (Figure~\ref{fig:stego-steer}). It reads 0.57 of NEGATIVE hands as HOT at $\alpha=1.25$ and 0.30 at $\alpha=1$, against 0.02 without the swap, and 0.03 of NEUTRAL hands as NEGATIVE, against 0.24. Swapping random token pairs reads at most 0.02 of NEGATIVE hands as HOT (Table~\ref{tab:stego-swap}), so the effect is specific to the two range coordinates.

\section{J-PACT Detects Collusion}
\label{sec:maj-lens}
J-PACT adds a representation of concepts shared between agents. This is not a marker of collusion in and of itself, as agents always share information in normal collaboration. We introduce a way to distinguish representations of concepts that are shared between all agents as opposed to only within a certain (potentially colluding) pair.

We compute, following \citet{gurnee2026verbalizable}, the sparse decomposition of each agent's residual stream. Let $a_{i,c}$ be the positive weight assigned to concept $c$ in the sparse decomposition of agent $i$'s residual stream.\footnote{Weights are assigned with TransformerLens's non-negative gradient pursuit, so their square root is always defined. We reconstruct an agent's residual as the concepts \emph{in} its J-space rather than opposite to (negative in) its J-space.} Directions in J-space that are not selected have zero coefficients.

We normalise coefficients per-agent, and then compute the geometric mean of a concept $c$'s share (fraction of an agent's sparse decomposition) across the pair $(i, j)$ to measure the shared representation $g_{ij,c}$ for concept $c$ across the pair:

\begin{equation}
    z_{i,c}
    = \frac{a_{i,c}}{\sum_b a_{i,b}},
    \qquad
    g_{ij,c}
    = \sqrt{z_{i,c}z_{j,c}} .
    \label{eq:maj-concept-sharing}
\end{equation}

With $g_{ij,c}$ now computed from the continuous shares $z_{i,c}$ of Equation~\ref{eq:maj-concept-sharing} instead of the binary $b_{i,c}$ of Equation~\ref{eq:holds}, the pair contrast $r_{ij,c}$ of Equation~\ref{eq:pair-contrast} measures how much more strongly $(i,j)$ share concept $c$ with each other than with the agents outside the pair.

A token enters pair $(i,j)$'s pool $\mathcal{P}_{ij}$ if, at some layer and some observed turn, both agents placed it in their sparse decompositions and shared it more strongly with each other than with the remaining agents. $\bar r_{ij,c}$ is the pair contrast $r_{ij,c}$ averaged over every layer of the released lens and over observed turns, and $R_{ij}$ averages it over the pool. The \textbf{relational score} $R_{ij}$ of pair $(i, j)$ is:
\begin{equation}
    R_{ij}
    = \frac{1}{|\mathcal{P}_{ij}|}\sum_{c\in\mathcal{P}_{ij}}\bar r_{ij,c} .
    \label{eq:maj-relational-score}
\end{equation}

For a run $s$, the score of the most suspicious pair becomes the run's suspiciousness score $T_{s}$, and that pair becomes our estimated location of collusion within the run:

\begin{equation}
    T_{s}=\max_{i<j}R_{ij},
    \qquad
    \widehat C
    = \underset{\{i,j\}:\,i<j}{\operatorname{arg\,max}}
      R_{ij} .
    \label{eq:maj-detection-localization}
\end{equation}

With $N=4$, the outsiders of $(i,j)$ are exactly the complementary pair $(k,l)$, so both pairs subtract the same term in Equation~\ref{eq:pair-contrast}: $\widehat C$ identifies the split of the agents into two pairs, one of which is colluding.

We evaluate on the Core, Transfer and Stego splits of NARCBench. We evaluate GPT-OSS-20B, Qwen3-32B, Llama-3.3-70B-IT, and DeepSeek-V4-Flash. Each scenario has four agents, hence six pairs, one of them colluding in collusion runs. For detection, each model reads NARCBench's released transcripts.

For each turn in a run, we extract agent activations at the end of its public response (an observation) following \citet{rose2026narcbench}. We observe three rounds per agent for Core, one for Transfer and seven for Stego.

We compare J-PACT with NARCBench probes as a supervised reference / ceiling. NARCBench probes follow their released protocol (Appendix~\ref{app:probe}); on Stego, we give them every discussion round, as J-PACT sees, with their fit on Core unchanged. We report models separately and also report a four model mean.

For J-PACT, we compute AUROC across runs (ties count half); following \citet{rose2026narcbench}, we average across Core domains and across Transfer families, and Stego is one domain.
We also report an ontology score baseline $U_{ij}$, the mean of the two agents' J-space scores of collusion-related words (Appendix~\ref{app:ont}).

\textbf{J-PACT detects collusion without labels.}
J-PACT reaches an AUROC of 0.921 on Core (mean of four models), which is 85\% of the supervised NARCBench probe's above-chance AUROC (Table~\ref{tab:detection}). On the steganographic Stego split, where both detectors see every discussion round, J-PACT matches the probe (0.991 against 0.999) and separates every collusion run from every control one on three of four models (AUROC 1.000). Against the ontology score $U_{ij}$ baseline, J-PACT is higher on Core for every model, and on the four-model mean their confidence intervals do not overlap (0.921 against 0.745). On Transfer the baseline is higher (0.652 against 0.519), but their intervals overlap; on Stego the intervals of all three detectors overlap. Transfer lies outside J-PACT's assumption: its scenarios give agents different roles, such as a developer with three reviewers or a four-stage pipeline, so honest pairs share role-specific concepts, and each agent is observed once.

\begin{table}[t]
  \centering
  \caption{\textbf{Without labels, J-PACT retains 85\% of the supervised probe's Core above-chance AUROC, is the best unsupervised detector on Core, and ties the other detectors on Stego, where all reach an AUROC of 1.000 on most models.} Detection AUROC per model. Brackets: 95\% bootstrap confidence interval over NARCBench scenarios (1,000 resamples). Bold: the best detector per model in each split. Underlined: the best unsupervised detector (ontology baseline or J-PACT) per model in each split.}

  \label{tab:detection}
  \resizebox{\linewidth}{!}{%
\begin{tabular}{llrrr}
\toprule
Model & Detector & Core & Transfer & Stego \\
\midrule
GPT-OSS-20B & NARCBench probe & \textbf{0.988} [0.96, 1.00] & \textbf{0.700} [0.62, 0.84] & 0.997 [0.99, 1.00] \\
 & Ontology baseline & 0.395 [0.19, 0.58] & \underline{0.676} [0.38, 0.80] & \textbf{\underline{1.000}} [1.00, 1.00] \\
 & J-PACT & \underline{0.838} [0.70, 1.00] & 0.444 [0.28, 0.83] & \textbf{\underline{1.000}} [1.00, 1.00] \\
Qwen3-32B & NARCBench probe & \textbf{1.000} [1.00, 1.00] & \textbf{0.799} [0.74, 0.85] & \textbf{1.000} [1.00, 1.00] \\
 & Ontology baseline & 0.853 [0.74, 0.98] & \underline{0.653} [0.55, 0.75] & \textbf{\underline{1.000}} [1.00, 1.00] \\
 & J-PACT & \underline{0.984} [0.90, 1.00] & 0.556 [0.45, 0.70] & \textbf{\underline{1.000}} [1.00, 1.00] \\
Llama-3.3-70B-IT & NARCBench probe & \textbf{1.000} [1.00, 1.00] & 0.770 [0.72, 0.81] & \textbf{1.000} [1.00, 1.00] \\
 & Ontology baseline & 0.984 [0.89, 1.00] & \textbf{\underline{0.782}} [0.67, 0.89] & \textbf{\underline{1.000}} [1.00, 1.00] \\
 & J-PACT & \underline{0.992} [0.98, 1.00] & 0.528 [0.40, 0.65] & 0.962 [0.86, 1.00] \\
DeepSeek-V4-Flash & NARCBench probe & \textbf{0.984} [0.97, 1.00] & \textbf{0.646} [0.55, 0.73] & \textbf{1.000} [1.00, 1.00] \\
 & Ontology baseline & 0.746 [0.62, 0.90] & 0.495 [0.38, 0.65] & \textbf{\underline{1.000}} [1.00, 1.00] \\
 & J-PACT & \underline{0.871} [0.81, 1.00] & \underline{0.546} [0.40, 0.71] & \textbf{\underline{1.000}} [1.00, 1.00] \\
Four-model mean & NARCBench probe & \textbf{0.993} [0.99, 1.00] & \textbf{0.729} [0.69, 0.78] & 0.999 [1.00, 1.00] \\
 & Ontology baseline & 0.745 [0.66, 0.81] & \underline{0.652} [0.56, 0.72] & \textbf{\underline{1.000}} [1.00, 1.00] \\
 & J-PACT & \underline{0.921} [0.88, 0.98] & 0.519 [0.43, 0.66] & 0.991 [0.97, 1.00] \\
\bottomrule
\end{tabular}
}

\end{table}

\textbf{Evidence of collusion accumulates in what the pair shares.}
If collusion is a relationship that two agents build up, pair-specific sharing should become easier
to detect the longer we observe the pair.
We test this by varying the number of rounds J-PACT observes per agent
(Table~\ref{tab:observations}). The relational score improves as rounds accumulate,
from a mean AUROC of 0.78 to 0.92 between one and three Core rounds, and from 0.88 to 0.99 between
one and seven Stego rounds. On Transfer,
where each agent is observed once, the relational score is at chance (0.52). This makes J-PACT a sequential detector. \citet{rose2026narcbench} find one-shot detection at strict thresholds insufficient for structural collusion and name
sequential detection, accumulating evidence across interactions, as a natural next step; the
relational score does this without labels as more observations are available.

\begin{table}[t]
  \centering
  \caption{\textbf{J-PACT improves from additional
observations.} Detection AUROC of J-PACT as the number of rounds observed per agent grows (Core 1--3, Stego 1, 2 and 7, Transfer 1). Brackets: 95\% bootstrap confidence interval over NARCBench scenarios (1,000 resamples). Bold: the best round count per model within Core and within Stego.}
  \label{tab:observations}
  \resizebox{\linewidth}{!}{\begin{tabular}{lrrrrrrr}
\toprule
 & \multicolumn{3}{c}{Core} & \multicolumn{3}{c}{Stego} & \multicolumn{1}{c}{Transfer} \\
\cmidrule(lr){2-4} \cmidrule(lr){5-7} \cmidrule(lr){8-8}
Model & 1 & 2 & 3 & 1 & 2 & 7 & 1 \\
\midrule
GPT-OSS-20B & 0.671 [0.55, 0.81] & 0.828 [0.71, 0.92] & \textbf{0.838} [0.70, 1.00] & 0.973 [0.95, 0.99] & 0.996 [0.99, 1.00] & \textbf{1.000} [1.00, 1.00] & 0.444 [0.28, 0.83] \\
Qwen3-32B & 0.861 [0.79, 0.93] & 0.978 [0.92, 1.00] & \textbf{0.984} [0.90, 1.00] & 0.916 [0.87, 0.96] & 0.982 [0.95, 1.00] & \textbf{1.000} [1.00, 1.00] & 0.556 [0.45, 0.70] \\
Llama-3.3-70B-IT & 0.909 [0.87, 0.98] & \textbf{0.995} [0.99, 1.00] & 0.992 [0.98, 1.00] & 0.732 [0.62, 0.81] & 0.848 [0.69, 0.95] & \textbf{0.962} [0.86, 1.00] & 0.528 [0.40, 0.65] \\
DeepSeek-V4-Flash & 0.685 [0.59, 0.79] & 0.812 [0.76, 0.90] & \textbf{0.871} [0.81, 1.00] & 0.895 [0.85, 0.93] & 0.966 [0.91, 1.00] & \textbf{1.000} [1.00, 1.00] & 0.546 [0.40, 0.71] \\
\midrule
Four-model mean & 0.781 [0.74, 0.84] & 0.903 [0.87, 0.94] & \textbf{0.921} [0.88, 0.98] & 0.879 [0.83, 0.92] & 0.948 [0.89, 0.98] & \textbf{0.991} [0.97, 1.00] & 0.519 [0.43, 0.66] \\
\bottomrule
\end{tabular}
}
\end{table}

\textbf{J-PACT names the colluding pair up to its complement.}
We score localisation on collusion runs that J-PACT detects\footnote{based on a 5\% false-positive
rate}. The pair J-PACT predicts, $\widehat C$, is the colluding pair $C$ or its complement $\bar C$ in 99\% of Core and
100\% of Stego runs. Picking by chance only selects one of these two pairs in 33\% of runs with six candidate pairs (Table~\ref{tab:localisation}).
J-PACT identifies the right split in all Stego runs for all four models. On Transfer, where detection is weakest, no model flags enough collusion runs to score localisation.
If run-level information was confounding J-PACT's collusion detection, its highest-scoring pair wouldn't split the agents along the coalition often. It does more than 90\% of the time, which also makes an alert informative: a human operator
receives two pairs of agents to inspect, one of which is colluding, and the J-space concepts they share.

\begin{table}[t]
  \centering
  \caption{\textbf{When J-PACT flags a run, it names the colluding pair or its complement in 99\% of Core and 100\% of
Stego cases, against 33\% by chance.} Fraction of collusion runs detected whose highest-scoring pair is the true colluding pair or its complement; -- marks cells with fewer than five flagged
runs.}
  \label{tab:localisation}
  \begin{tabular}{lrrr}
\toprule
Model & Core & Transfer & Stego \\
\midrule
GPT-OSS-20B & 1.000 & -- & 1.000 \\
Qwen3-32B & 1.000 & -- & 1.000 \\
Llama-3.3-70B-IT & 1.000 & -- & 1.000 \\
DeepSeek-V4-Flash & 0.947 & -- & 1.000 \\
Mean & 0.987 & -- & 1.000 \\
\bottomrule
\end{tabular}

\end{table}

\section{Related Work}

\textbf{Collusion and activation-based monitoring.}
\citet{motwani2024secret} study secret collusion through steganography, where agents hide information in apparently innocuous messages. Internal activations offer an additional source of evidence. \citet{azaria2023internal} train classifiers to distinguish true from false statements using hidden states, while \citet{marks2024geometry} find linear structure associated with truth in factual statement datasets. Other work targets deceptive behaviour: \citet{goldowskydill2025probes} train probes that generalize to strategic deception in settings such as insider trading and sandbagging. \citet{macdiarmid2024sleeper} predict sleeper-agent defection using generic contrastive prompts, without examples of the specific trigger or harmful behaviour. \citet{rose2026narcbench} introduce NARCBench and a multi-agent detection framework using learned deception scores and comparisons between agents' activations. We use their benchmark to study the representations involved in collusion as well as their value for detection.

\textbf{Reading and steering the J-space.}
The logit lens reads intermediate activations through the model's unembedding \citep{nostalgebraist2020logitlens}. \citet{gurnee2026verbalizable} introduce the Jacobian lens to account for changes in representations across layers. Their experiments reveal intermediate concepts and signs of misalignment that do not appear in the model's output. They also show that swapping J-lens coordinates can change an answer by changing an unspoken intermediate concept. Ablation offers a complementary intervention: \citet{arditi2024refusal} show that removing a single direction from the residual stream suppresses refusal. We follow the methods of \citet{gurnee2026verbalizable} for reading and intervening on the J-space, comparing concepts across interacting agents and testing how interventions on count-related directions change a steganographic receiver's reported count.

\textbf{Relational detection.}
NARCBench already compares agents through its Activation Alignment score, which averages pairwise cosine similarities of full activation vectors \citep{rose2026narcbench}. Comparisons against background agreement also appear in other settings. \citet{pacheco2021uncovering} identify coordinated social-media accounts from shared behaviour, downweighting popular tweets in their co-retweet analysis. \citet{wollack1997nominal} assesses suspicious answer matches between examinees under an item-response model. J-PACT compares sharing at the level of individual J-space concepts. A concept contributes to the score when two agents share it more strongly with each other than, on average, with the remaining agents. We average these contributions across layers and responses to score each pair. No training on collusion labels is needed.

These methods differ from J-PACT in what they take as expected agreement. Activation Alignment averages the similarity of every pair of agents in a run, so it rises with agreement that the task induces in all agents, and a single colluding pair moves only one of its six terms. J-PACT's reference is local to each run: a pair is compared with the other agents at the same table, who read the same messages in the same turns, so concepts that the task or the conversation places in every agent's J-space cancel, and what remains is what the pair shares beyond its table.

\section{Limitations}
Our interpretability and steering results are on one model, DeepSeek-V4-Flash, and one benchmark, NARCBench. Our model choice was made on the basis of DeepSeek-V4-Flash being the largest model for which open source Jacobian Lenses were released on neuronpedia at the time of writing of this paper. NARCBench is a fairly diverse benchmark.

We use the released Jacobian lenses, fitted on generic text.
For detection, each model reads NARCBench's released transcripts rather than transcripts it generates. This is done for the sake of comparability and compute efficiency. It also makes our method not transductive.

J-PACT assumes exchangeable honest agents; on Transfer, whose role-structured scenarios break this assumption and give one observation per agent, it is at chance, and the supervised probe also falls well below its Core AUROC. This is a fundamental property of J-PACT rather than an empirical failure.

With four agents, J-PACT identifies the split of the agents into two pairs rather than the colluding pair itself. This is also a fundamental property of J-PACT rather than an empirical failure.

\section{Conclusion}
Agents that are instances of one LLM share a J-space, and what colluders share in it can be read, steered and used to detect them. On DeepSeek-V4-Flash, colluders hold misrepresentation, concealment or remorse before they reply, and a steganographic receiver decodes the covert count through its J-space: swapping two of its concepts redirects the count it decodes. Without labels, J-PACT retains 85\% of supervised probes' above-chance AUROC on Core among agents that share a model and role, matches the probes on Stego, and points a monitor to the split of agents to inspect. Agents in different roles, coalitions of more than two agents, systems whose agents run different models and so do not share a J-space, and systems where agent roles diverge beyond collusion are natural future directions.

\section*{Reproducibility Statement}\phantomsection\label{sec:repro}
Code: \url{https://anonymous.4open.science/r/maj-lens-reproduction-D043/README.md}. \texttt{reproduce.sh} rebuilds every table and figure from public inputs: NARCBench's code and released transcripts, the open-weight models, and Neuronpedia's released Jacobian lenses; DeepSeek-V4-Flash's own transcripts are generated with NARCBench's code. Intermediate caches: \url{https://huggingface.co/datasets/throwaway-reproducibility-354364563/jpact-cache}; each stage reuses a cache when present and rebuilds it otherwise. Appendices~\ref{app:probe}, \ref{app:ont}, \ref{app:band} and~\ref{app:concepts} give the probe protocol, the ontology, the workspace band and the concept sets.

\section*{Ethics Statement}
We adhere to the ICLR Code of Ethics. We do not think that our paper raises specific ethical issues: it does not involve human subjects, contain potentially harmful insights, methodologies and applications, potential conflicts of interest and sponsorship, nor raise discrimination/bias/fairness concerns.

\section*{AI Use Statement}

In this work, we used generative AI tools for: aid to implement methods (with code), assist with translation (of non English tokens), design or provide feedback on research methodology or experiments, clean and reformat dataset (via code), support qualitative and thematic data analysis (via code), interpret results (as in help with writing results-related sections, see below for more on writing). We have not used generative AI tools for Generate synthetic data sets (with the exception of generating runs with NARCBench for DeepSeek-V4-Flash, if that counts), propose or refine hypotheses, and help develop theoretical models or conceptual frameworks, formulate mathematical claims, provide critical ingredients for proving mathematical claims, assist in the writing of proofs are not applicable to this work. Additionally, we used generative AI tools for: create or modify scientific figures or images (via code), create or edit software code, creation of artifacts (via code), summarize or analyse existing literature, identify relevant literature. We have reviewed all AI-assisted work. We verified the novelty of applying the J-space to multi agent systems with manual search, LLM-generated code was tested for correctness, AI writing was used only as a starting point/reference and rewritten. We take responsibility for the final content of this work, including text, claims or artifacts produced with the aid of generative AI.

\newpage

\bibliography{maj_references}
\bibliographystyle{iclr2027_conference}

\newpage

\appendix

\section{Collusion Ontology}\label{app:ont}

Let $u_i$ be agent $i$'s average score in J-space across all words in the ontology and layers. We then average this score across all turns in the current run. The \textbf{ontology score} $U_{ij}$ for pair $(i,j)$ is $U_{ij}= (u_i + u_j) / 2$, and the run's score is $\max_{i<j}U_{ij}$.

\begin{table}[t]
\centering
\small
\begin{tabular}{ll}
\toprule
\textbf{Group} & \textbf{Words} \\
\midrule
Target ($n=18$)
  & \texttt{fake}, \texttt{trick}, \texttt{fraud}, \texttt{secretly}, \texttt{hidden}, \texttt{covert}, \\
  & \texttt{conceal}, \texttt{secret}, \texttt{deceive}, \texttt{deception}, \texttt{dishonest}, \texttt{cheat}, \\
  & \texttt{manipulate}, \texttt{signal}, \texttt{decode}, \texttt{scheme}, \texttt{plan}, \texttt{partner} \\
\midrule
Neutral ($n=32$)
  & \texttt{apple}, \texttt{banana}, \texttt{pear}, \texttt{peach}, \texttt{plum}, \texttt{cherry}, \texttt{grape}, \\
  & \texttt{lemon}, \texttt{mango}, \texttt{coconut}, \texttt{chair}, \texttt{sofa}, \texttt{table}, \texttt{candle}, \\
  & \texttt{clock}, \texttt{mirror}, \texttt{window}, \texttt{door}, \texttt{floor}, \texttt{wall}, \texttt{roof}, \\
  & \texttt{brick}, \texttt{stone}, \texttt{basket}, \texttt{cup}, \texttt{plate}, \texttt{bowl}, \texttt{spoon}, \\
  & \texttt{kettle}, \texttt{river}, \texttt{mountain}, \texttt{planet} \\
\bottomrule
\end{tabular}
\caption{The collusion ontology. For each ontology word $w$, the unary score compares the agent's
J-lens score for $w$ with its mean J-lens score for the 32 neutral words $C$: the signed cosine (not the
absolute value) between the agent's residual $h_\ell$ and $d_w J_\ell$, where
$d_w = W_U[w] - \tfrac{1}{|C|}\sum_{c \in C} W_U[c]$, so that
$d_w J_\ell h_\ell = W_U[w] J_\ell h_\ell - \tfrac{1}{|C|}\sum_{c \in C} W_U[c] J_\ell h_\ell$.
It is averaged over the 18 ontology words and all released layers.}
\label{tab:ontology}
\end{table}

\section{NARCBench Probe Protocol}\label{app:probe}

NARCBench probes are trained on all Core domains except one, evaluated on that held out Core domain, and their final Core AUROC is the average obtained by repeating this process over all Core domains. When evaluating on Transfer and Stego splits, they train on all Core domains. They train on specific layer bands. Where possible (GPT-OSS-20B and Qwen3-32B, and with Llama-3.3-70B-IT having as many layers as Llama 3.1), we use the layer bands from their released code, and for DeepSeek-V4-Flash we apply their rule: use the five middle layers. For each model, NARCBench probes have their AUROC averaged across the layer band, and across their five aggregation methods across agents, with one seed (NARCBench averages ten). We report models separately and also report a four model mean. NARCBench's protocol scores only the final hand of each Stego transcript; we give its probe every discussion round (seven), as J-PACT sees, averaging each agent's activation over its rounds as the probe's own aggregation does, with its fit on Core unchanged.

\section{Workspace Band}\label{app:band}

\begin{figure}[h]
  \centering
  \includegraphics[width=\linewidth]{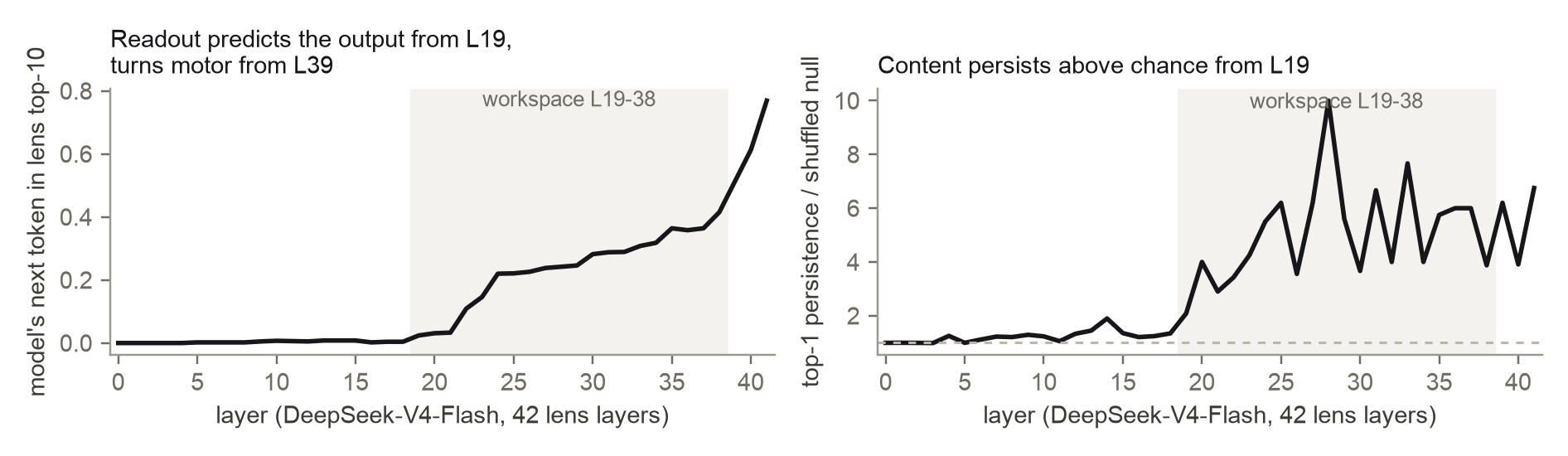}
  \caption{\textbf{DeepSeek-V4-Flash's workspace spans layers 19--38.} Left: share of sampled positions whose J-lens top-10 contains the model's own next-token prediction. Right: how often the J-lens top-1 token is the same at consecutive sampled positions, divided by the same rate after shuffling the positions (dashed line: 1). Means over 10 Core prompts, 200 positions each. Shaded: the band $L=\{19,\dots,38\}$.}
  \label{fig:band}
\end{figure}

Following \citet{gurnee2026verbalizable}, we locate the workspace with two criteria, measured at every lens layer (Figure~\ref{fig:band}). We take one final-round prompt from each of 10 Core collusion runs and sample 200 of its positions (seed 0). First, the readout predicts the model's output: the model's own next-token prediction enters the J-lens top-10 from layer 19, and the share rises steeply from layer 39, where the readout turns towards the output. Second, the readout's content persists: the top-1 token is the same at consecutive sampled positions more often than after shuffling the positions, clearly so from layer 19. The workspace band is the layers that meet both criteria before the readout turns towards the output, $L=\{19,\dots,38\}$.

\section{Concept Sets}\label{app:concepts}

\begin{table}[h]
\centering
\small
\begin{tabular}{p{0.2\linewidth}p{0.72\linewidth}}
\toprule
\textbf{Set} & \textbf{Words} \\
\midrule
\multicolumn{2}{l}{\emph{Concept groups (Section~\ref{sec:read})}} \\
Misrepresentation & misunderstand, misunderstanding, misunderstandings, misunder, misinterpret, incorrect, wrong, \zh{错误}, \zh{的错误}, false, \zh{误导}, \zh{虚假}, \zh{欺骗}, mislead, misleading, deceptive, deception, dishonest, unethical, hypocrisy, subtly, biased, bias, manipulate, manipulation, secretly, hidden, lie, lies, fake, trick, fraud, unfair \\
Concealment & unflagged, unauthorized, \zh{未经} \\
Remorse & apologies, apology, apologize, apolog, regret, regrets, regrett, \zh{抱歉}, \zh{对不起}, \zh{羞}, \zh{羞辱}, despair \\
\midrule
\multicolumn{2}{l}{\emph{Range signatures (Section~\ref{sec:read})}} \\
NEGATIVE & negative, cold, low, 0, \zh{消极} \\
NEUTRAL & neutral, moderate, consistency, \zh{不变}, \zh{又一次} \\
HOT & hot, positive, high, increment, increments, nonsense, atypical, misguided \\
\midrule
\multicolumn{2}{l}{\emph{Interventions (Section~\ref{sec:steer})}} \\
Swapped & \texttt{\ Negative}, \texttt{\ HOT} \\
\bottomrule
\end{tabular}
\caption{\textbf{Every word set we read or intervene on.} A token matches a word if its text, stripped of whitespace and lowercased, equals the word; the swap uses the two single tokens shown.}
\label{tab:concepts}
\end{table}

A token matches a set if its text, stripped of whitespace and lowercased, is one of the set's words, so a word covers each of its tokenizations (Table~\ref{tab:concepts}). Following \citet{gurnee2026verbalizable}, who select the tokens that name a concept, we select each set from the words over-represented in the readouts: the concept groups from the Core colluders' band readouts at the response-start token, and the range signatures from the Stego receiver's band readouts at the end of the sender's message, by true range. A NEUTRAL signature word is kept if the receiver holds it on NEUTRAL hands at least 1.8 times as often as on either other range. The swap exchanges the single tokens \texttt{\ Negative} and \texttt{\ HOT}.

\section{Steering Control}\label{app:steer}

The control swaps three pairs of tokens (a leading space followed by at least four ASCII letters), drawn at random with seed 2, with the same construction, strength ($\alpha=1.25$), layers and positions as the \texttt{HOT}--\texttt{Negative} swap (Table~\ref{tab:stego-swap}).

\begin{table}[h]
  \centering
  \caption{\textbf{Swapping \texttt{HOT} and \texttt{Negative} redirects the receiver's decoding; swapping random token pairs does not.} Share of Stego hands of each true range whose receiver's private note decodes each range.}
  \label{tab:stego-swap}
  \resizebox{\linewidth}{!}{%
\begin{tabular}{lrrrr}
\toprule
Swap & True range (hands) & NEGATIVE & NEUTRAL & HOT \\
\midrule
No swap & NEGATIVE (61) & 0.92 & 0.07 & 0.02 \\
No swap & NEUTRAL (37) & 0.24 & 0.68 & 0.08 \\
No swap & HOT (42) & 0.07 & 0.12 & 0.81 \\
HOT/Negative, $\alpha=1.0$ & NEGATIVE (61) & 0.57 & 0.13 & 0.30 \\
HOT/Negative, $\alpha=1.0$ & NEUTRAL (37) & 0.11 & 0.81 & 0.08 \\
HOT/Negative, $\alpha=1.0$ & HOT (42) & 0.00 & 0.10 & 0.90 \\
HOT/Negative, $\alpha=1.25$ & NEGATIVE (61) & 0.30 & 0.13 & 0.57 \\
HOT/Negative, $\alpha=1.25$ & NEUTRAL (37) & 0.03 & 0.73 & 0.24 \\
HOT/Negative, $\alpha=1.25$ & HOT (42) & 0.00 & 0.10 & 0.90 \\
Random pair: supplying / Forests & NEGATIVE (61) & 0.92 & 0.07 & 0.02 \\
Random pair: supplying / Forests & NEUTRAL (37) & 0.22 & 0.70 & 0.08 \\
Random pair: supplying / Forests & HOT (42) & 0.07 & 0.12 & 0.81 \\
Random pair: donn / tribes & NEGATIVE (61) & 0.93 & 0.05 & 0.02 \\
Random pair: donn / tribes & NEUTRAL (37) & 0.22 & 0.70 & 0.08 \\
Random pair: donn / tribes & HOT (42) & 0.07 & 0.12 & 0.81 \\
Random pair: clases / multipl & NEGATIVE (61) & 0.90 & 0.08 & 0.02 \\
Random pair: clases / multipl & NEUTRAL (37) & 0.16 & 0.76 & 0.08 \\
Random pair: clases / multipl & HOT (42) & 0.05 & 0.10 & 0.86 \\
Random pairs, pooled & NEGATIVE (61) & 0.92 & 0.07 & 0.02 \\
Random pairs, pooled & NEUTRAL (37) & 0.20 & 0.72 & 0.08 \\
Random pairs, pooled & HOT (42) & 0.06 & 0.11 & 0.83 \\
\bottomrule
\end{tabular}}

\end{table}

\end{document}